\documentclass[lettersize,journal]{IEEEtran}
\usepackage{lmodern}  
\usepackage[T1]{fontenc}
\usepackage[utf8]{inputenc}  
\usepackage{amsmath,amsfonts}
\usepackage{algorithmic}
\usepackage{algorithm}
\usepackage{array}
\usepackage[caption=false,font=normalsize,labelfont=sf,textfont=sf]{subfig}
\usepackage{textcomp}
\usepackage{stfloats}
\usepackage{flushend}
\usepackage{url}
\usepackage{verbatim}
\usepackage{graphicx}
\usepackage{cite}
\usepackage{booktabs}
\begin{document}

\title{Reinforcement Learning on Cost-Constrained Quadrupedal Hardware}

\author{Javier C. Weddington, Bence P. Ölveczky, Stephen A. Baccus
\thanks{J. C. Weddington is with the Neurosciences Interdepartmental Program, Stanford University, Stanford, CA 94305 USA (e-mail: javiercw@stanford.edu).
B. P. Ölveczky is with the Department of Organismic and Evolutionary Biology, Harvard University, Cambridge, MA 02138 USA.
S. A. Baccus is with the Department of Neurobiology, Stanford University School of Medicine, Stanford, CA 94305 USA.}
\thanks{Code and models: \texttt{github.com/baccuslab/SpotDMouse}\\(directory \texttt{P2-Terrain\_Challenge}).}}


\maketitle

\begin{abstract}
Deploying learned control policies on low-cost robotic platforms introduces transport latencies and noisy motor feedback that systematically widens the sim-to-real gap. The chasm of simulation to deployment in hardware lies in the delay of the actuator reaching the commanded position. On platforms such as the Mini Pupper 2, a measured $ > $50 ms transport delay transforms the locomotion task from a standard Markov decision process into a partially observable one. In this paper, we take a biologically inspired approach of handling noisy and delayed feedback to close the sim-to-real gap, thereby expanding the capability of reinforcement learning on cost-constrained hardware. Using a low-cost quadrupedal hardware platform, we find that using a forward model of the average actuator delay, paired with a time-aware neural network results in robust locomotion. Additionally, our time-aware neural network learned a central pattern generator (CPG): a self-sustaining rhythmic gait that is robust to +320 ms latency perturbations, mirroring the CPGs found in the spinal cords of vertebrates. We posit that temporal self-organization may be a general strategy for cost-constrained locomotion.
\end{abstract}


\section{Introduction}
\label{sec:introduction}

\IEEEPARstart{R}{einforcement} Learning (RL) on quadrupedal walking has made remarkable strides. Demonstrations of RL in quadrupeds show robustness to a wide array of adversarial disturbances, showing promise in applications for legged robots in industry. However, replicating these demos on affordable hardware remains out of reach for most. Standard quadrupeds use quasi-direct-drive or elastic actuators with low latency and high control frequencies~\cite{rudin2022, lee2020, miki2022}. This trivializes the incorporation of delay in training and deployment, since the maximum amount of delay fits within one iteration of the control loop. At \$10K--\$150K, purchasing power buys machine fidelity on dynamic robotic control tasks, introducing two problems for robotics practitioners and companies: high fixed cost and high barrier to entry. In this paper, we explore methods to overcome low-cost constraints for RL quadrupedal locomotion.

\begin{figure}[!t]
\centering
\includegraphics[width=3.4in]{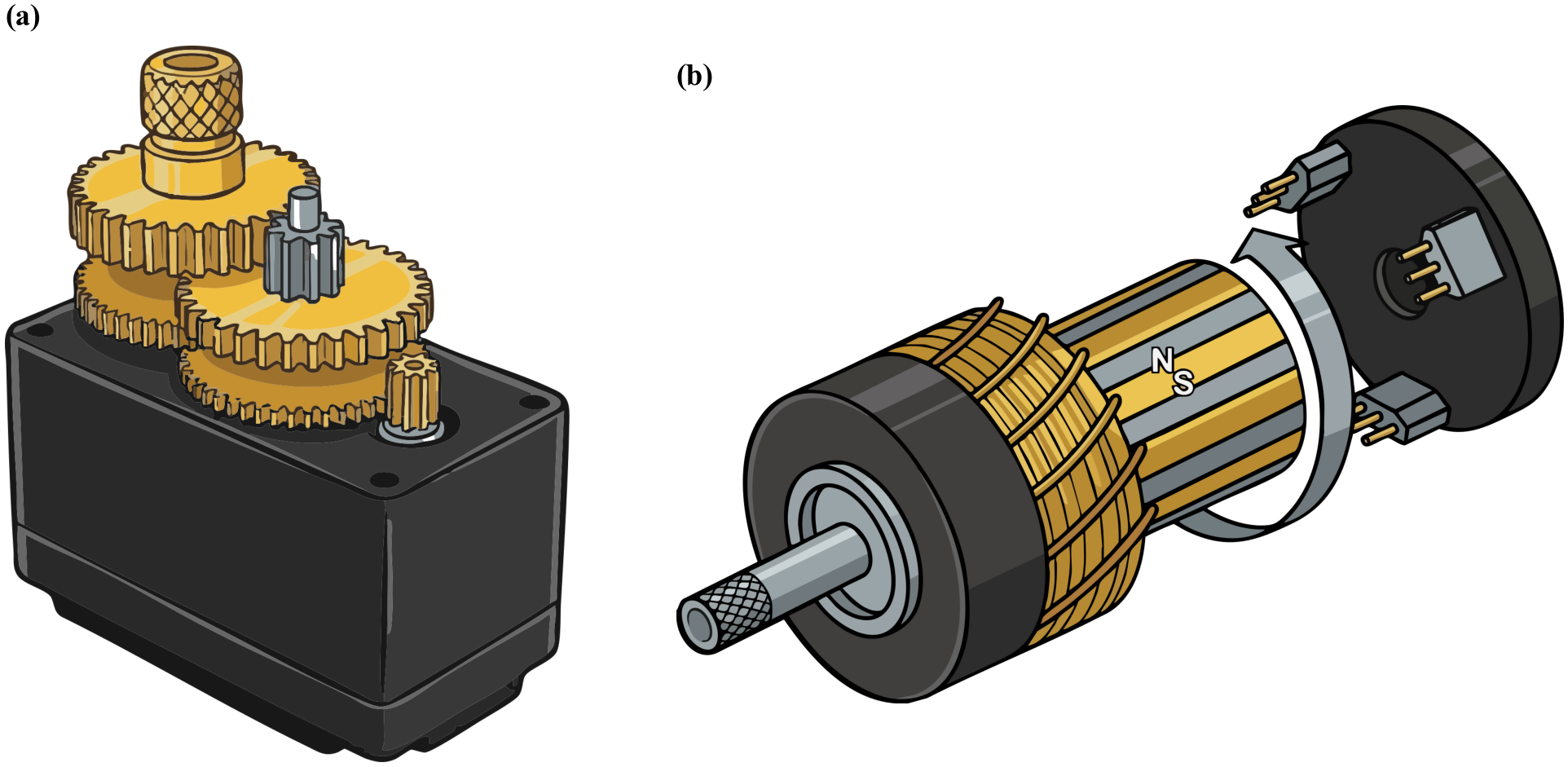}
\caption{Comparison of actuator architectures. (a)~Brushed DC position servo (Mini Pupper 2): \$8, position-only feedback, 76\,ms communication delay via ESP32 serial path. (b)~Brushless quasi-direct-drive motor (research platforms): \$200--\$500, torque/velocity/position feedback, $<$5\,ms latency.}
\label{fig:brushed_vs_brushless}
\end{figure}

Explicitly modeling actuator latency is mandatory for sim-to-real transfer~\cite{tan2018}, but prior methods target high-frequency brushless systems. Recent work on cost-constrained brushed-servo platforms exposes severe non-linearities and sluggish response times~\cite{mhaske2025}, and delay-resolved RL approaches incorporate action history or disturbance-aware representations to handle the resulting partial observability~\cite{malmir2023}. Yet few studies evaluate whether different neural architectures intrinsically manage the partially observability created by extreme ($>$50\,ms) hardware delays, or whether engineering intervention is required instead.

Core differences between expensive and cheap actuators are that brushed motors are less reproducible and introduce significantly higher delays than brushless motors (Fig. 1, Fig. 2). The Mini Pupper 2 is a \$300 quadruped built around brushed DC position servos that communicate through an ESP32 microcontroller at a measured 76\,ms transport delay and only provides position feedback. This delay regime is not unique to inexpensive hardware, as biological sensorimotor systems operate under similar constraints with  delays of 100--160\,ms between motor commands and resulting sensory feedback~\cite{rasman2021, kuo2005, vanderkooij1999}. Inexpensive hardware, like biological systems, must also contend with noisy and incomplete state information, and have limited corrective bandwidth. Thus, simple feedback control becomes ineffective when time delays are long. By the time sensory signals reach the controller, the information is dated. By the time motor commands reach the actuators, they may be inappropriate for the new joint state~\cite{more2010}. Taking inspiration from biology, we framed the RL problem as a partially observable Markov decision process (POMDP): because the delay at each timestep varies, the network must evaluate its own temporal state in order to maintain a robust locomotion gait pattern. This suggests that static networks are not sufficient on their own, and that time-aware networks are a candidate solution for low-cost hardware. It is unknown, however, whether this more complex architecture is worthwhile or whether hand-engineered compensation for a simpler network is sufficient.

\begin{figure*}[!t]
\centering
\includegraphics[width=7.0in]{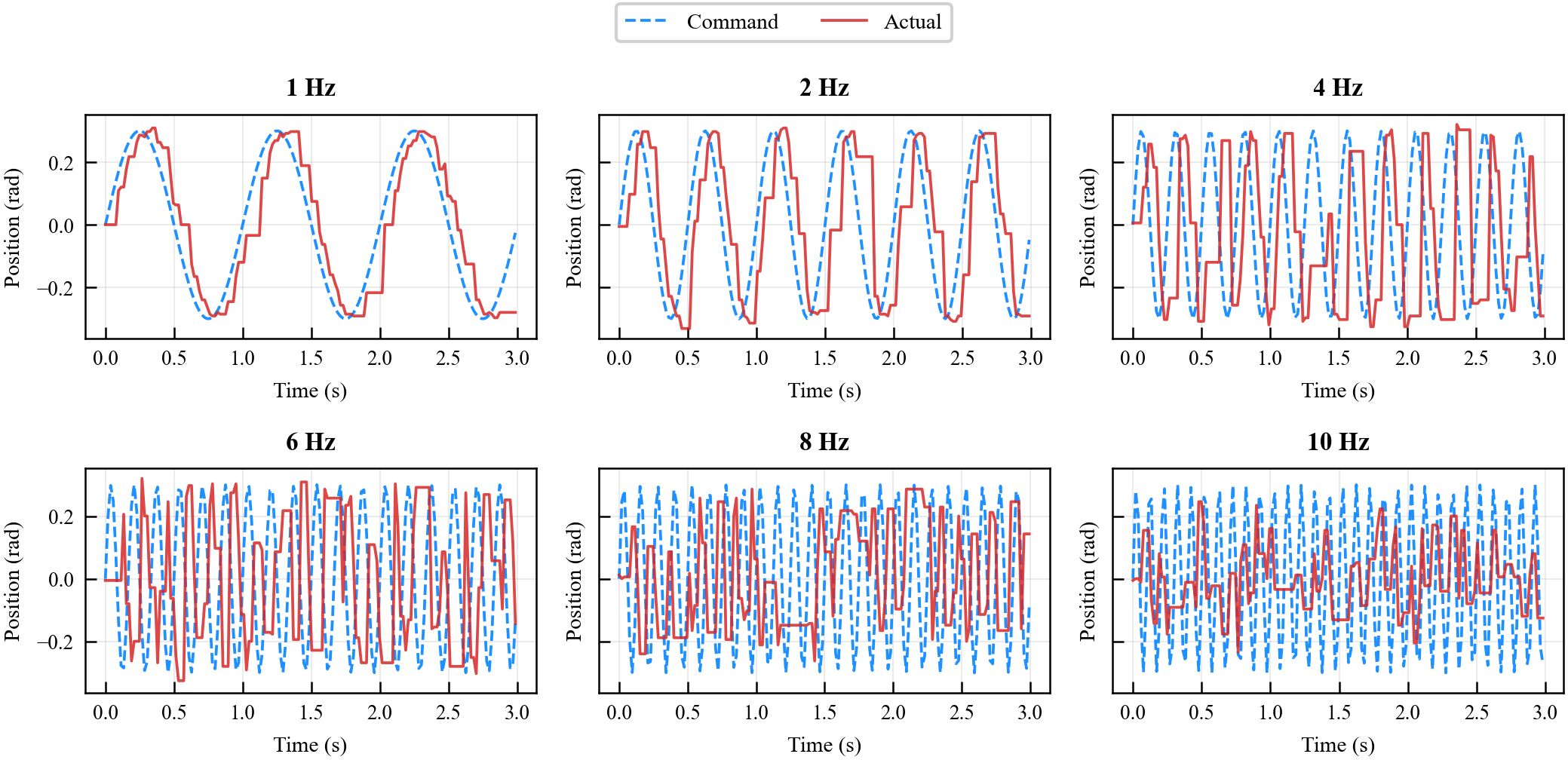}
\caption{Identification of actuator transport delay and tracking bandwidth. Real servo frequency response: sinusoidal position commands (blue dashed) vs.\ measured servo position (red solid) at 1--10\,Hz. Cross-correlation reveals a constant 76\,ms transport delay at 1--6\,Hz. Tracking degrades above 3\,Hz.}
\label{fig:delay_identification}
\end{figure*}

\subsection{Problem Statement}

On cost-constrained hardware, is the sim-to-real bottleneck best addressed through network architecture or through training and deployment engineering? Specifically, is it cheaper to build a model-based observation bridge such as synthetic proportional derivative (PD) feedback for a simple feedforward policy, or to train a temporal network that learns the delay and missing-state dynamics internally?

Let the state at time $t$ be $s_t$. If the system has an expected actuation delay of $k$ time steps, the action issued at time $t$ does not take effect until $t + k$. To maintain the Markov property, the augmented state must contain the buffer of actions previously issued but not yet executed:
\begin{equation}
\label{eq:pomdp}
\tilde{s}_t = \bigl(s_t,\; a_{t-1},\; a_{t-2},\; \ldots,\; a_{t-k}\bigr).
\end{equation}
For hardware where $s_t$ is itself only partially observable (no velocity, no torque, quantized position with stochastic delay $k$), the system can be formalized as a POMDP with observation space $\Omega$ and a stochastic delay buffer embedded in the underlying state~\cite{bouteiller2020, walsh2009, katsikopoulos2003}.

The POMDP framework explains multiple phenomena that arise when deploying reinforcement learning policies on low-cost locomotion hardware . These include why feedforward policies require elaborate deployment engineering, why certain recurrent architectures succeed while others fail, and why closing the observation loop can actively degrade performance on position-servo hardware.

\section{Methods}
\label{sec:methods}

\subsection{Training}

All policies were trained in NVIDIA Isaac Lab (IsaacSim) using the \texttt{rsl\_rl} proximal policy optimization (PPO) implementation~\cite{rudin2022} on a single NVIDIA RTX 4090 (24\,GiB) with 4098 parallel environments. The actuator model is a calibrated \texttt{DelayedPDActuatorCfg} with parameters matched to real servo delay and amplitude performance identified in Figure 2:
\begin{itemize}
\item Stiffness $K_p = 70$, damping $K_d = 1.2$
\item Friction $= 0.03$, armature $= 0.005$
\item Transport delay: 33--43 physics steps at 500\,Hz
\item Effort limit $= 5.0$\,Nm
\item Velocity limit overridden to 10.5\,rad/s via \texttt{velocity\_limit\_sim} 
\end{itemize}

The observation vector is 60-dimensional: base linear velocity~[3], base angular velocity~[3], projected gravity~[3], velocity commands~[3], relative joint positions~[12], joint velocities~[12], joint efforts~[12], and previous actions~[12]. Action scale is 0.5. We used tightened reward shaping tracking standard deviations ($\sigma_\text{lin} = 0.1$, $\sigma_\text{ang} = 0.2$) and penalty weights: foot slip $= -0.5$, joint position $= -0.7$, joint torques $= -5{\times}10^{-4}$, action smoothness $= -1.0$, base orientation $= -5.0$. 

\subsection{Network Architectures}

All architectures share a common MLP head: actor layers [512, 256, 128] and critic layers [512, 256, 128] with ELU activation. PPO hyperparameters are identical: learning rate $= 1{\times}10^{-3}$ (adaptive), $\gamma = 0.99$, $\lambda = 0.95$, clip ratio $= 0.2$, entropy coefficient $= 0.005$, 5 learning epochs per update, desired KL $= 0.01$, 24 steps per environment.


For temporal architectures, a recurrent or attention module maps observations to $\mathbb{R}^{128}$ before the MLP head.

\begin{table}[!t]
\caption{Architecture Comparison Under 76\,ms Transport Delay\label{tab:comparison}}
\centering\small
\begin{tabular}{@{}lcccc@{}}
\toprule
 & MLP & LSTM & GRU & Transf. \\
\midrule
Reward (40K iter)    & 271$^\dagger$ & 271  & 72   & $-$85.8 \\
Action noise std     & 0.07$^\dagger$& 0.07 & 0.25 & 1.13 \\
Curriculum           & Yes           & No   & N/A  & N/A  \\
GPU mem.\ (GiB)     & ${\sim}$12    & 12--24 & 12--24 & ${\sim}$24 \\
Steps/s              & 30K   & 15--20K & 15--20K & 5.5K \\
HW deploy.\ params  & 6+            & 1    & ---  & ---  \\
HW sensor obs        & 15/60  & 6/60  & --- & ---  \\
Terrain gen.         &No      & Yes  & ---  & ---  \\
\bottomrule
\multicolumn{5}{@{}p{\columnwidth}@{}}{\scriptsize $^\dagger$After curriculum. GRU/Transformer did not converge. HW deploy. params: hand-tuned deployment parameters (see Section II-C). HW sensor obs: real-sensor dimensions out of 60. Terrain gen.: gait generalizes to unseen terrain without retraining.}
\end{tabular}
\end{table}


All architectures (MLP, LSTM, GRU, Transformer) were first trained under identical reward and environment settings without curriculum. Under these conditions, only the LSTM converged to a stable gait in simulation. The MLP required a multi-stage curriculum; progressively increasing delay, penalty weights, and action scale to achieve multi-directional command tracking. The GRU and Transformer failed to converge under either condition within our training budget. The GRU's failure suggests that the task requires delay \emph{compensation} via specific architectural capacity (the cell state linear pathway), not merely oscillation capacity. If any oscillator sufficed, we would expect the GRU to converge. The Transformer's failure may not be purely computational. For a task where the relevant temporal dependency is a fixed $k$-step delay, the LSTM's cell state provides a shift-register structure. The Transformer must \textit{learn} which time steps matter, whereas carrying the delayed action forward is built into the LSTM's shift-register cell state as an inductive bias. Recent work has shown that Transformers can match or exceed LSTM performance in humanoid locomotion, but with four A100 GPUs, thousands of parallel environments, and a two-stage teacher–student pipeline that distills a privileged fully-observable policy into the deployed controller ~\cite{radosavovic2024}; the marginal gain (${\sim}$7\%) of the Transformer over the LSTM does not justify the compute cost for resource-constrained deployment.

\begin{figure}[!b]
\centering
\includegraphics[width=3.4in]{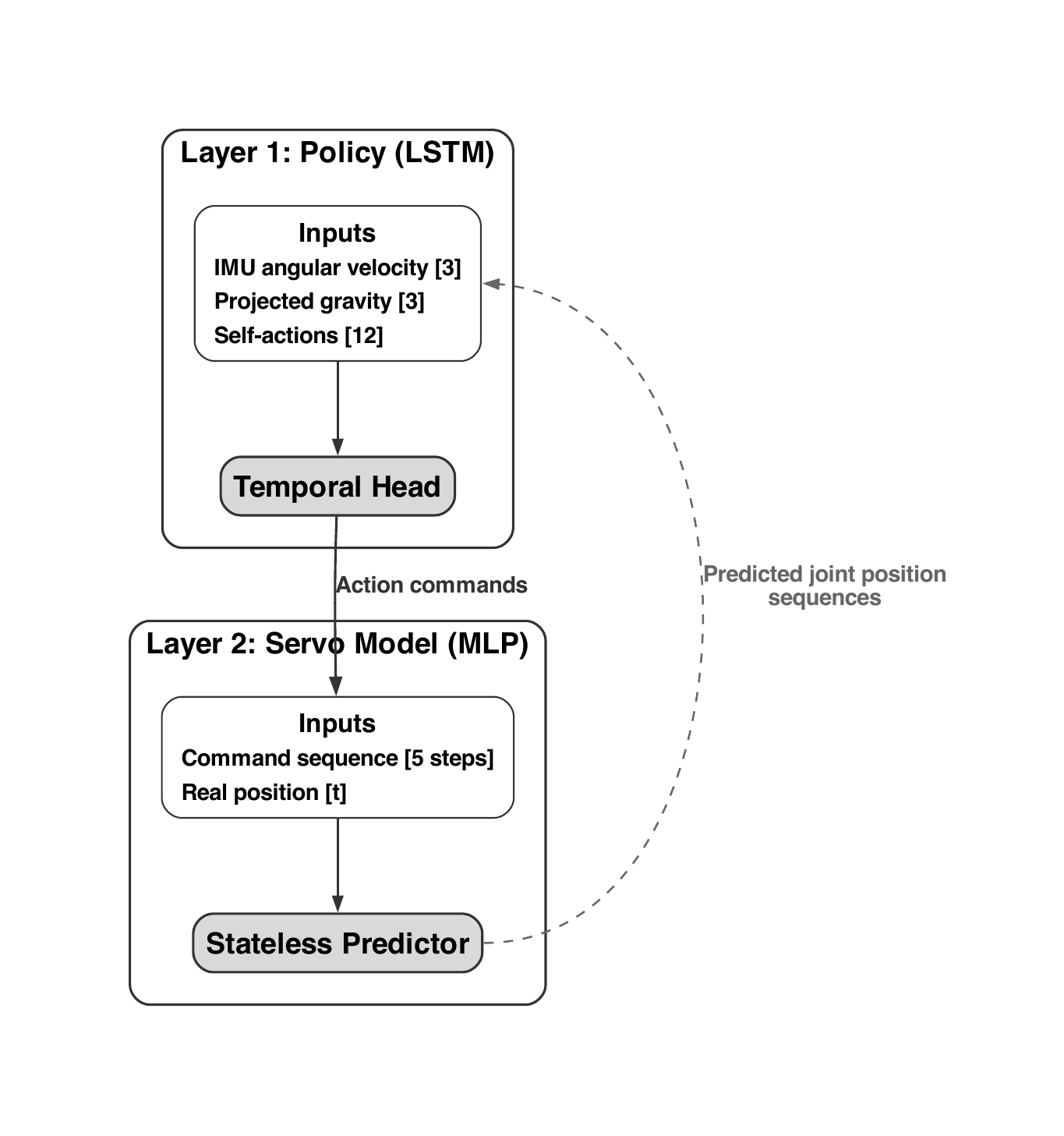}
\caption{Two-layer deployment architecture. The LSTM policy (Layer~1) handles long-horizon temporal tasks of delay compensation and gait generation, while the MLP servo model (Layer~2) provides short-horizon predicted joint positions via autoregressive rollout. The dashed arrow indicates the feedback path replacing dead hardware readings with learned servo dynamics.}
\label{fig:deployment_arch}
\end{figure}

\subsection{Deployment}
\label{sec:deployment}

Hardware deployment poses two distinct problems. First, low-cost position servos provide only position feedback: the servo reports joint velocity and effort as $\approx 0$. A policy trained with delay-free PD actuator dynamics is therefore missing observation channels it was trained to use. Overcoming this requires reconstructing the missing observations, which we address below. Second, each command takes 76 ms to reach the commanded position at the motor. When commands are issued faster than the servo can settle, above ~3\,Hz, tracking becomes  stifled (Fig. 2). Thus, reducing the command sampling rate allows the cost-constrained robot to meet the command in time for the next, which reduces compounding position errors. 

\textbf{MLP deployment} requires a synthetic PD observer that reconstructs the hidden state variables to execute the stable, learned gait:
\begin{enumerate}
\item Policy action enters a 5-step delay buffer (200\,ms at 25\,Hz control).
\item PD dynamics are simulated: $\tau = K_p(q_\text{target} - q) - K_d \dot{q}$, with 4 substeps per policy step (matching 500\,Hz physics / 50\,Hz policy in simulation).
\item Synthetic observations (joint position, velocity, effort) replace corrupt or noisy hardware readings.
\end{enumerate}
Of 60 observation dimensions, only 15 come from real sensors (25\%); 36 are synthetic (60\%), and 3 are heuristic (base linear velocity $= 0.7 \times$ command). Six or more parameters require tuning: $K_p$, $K_d$, inertia, delay steps, substeps, effort limit, bias scale, and hardware scale.

\textbf{LSTM deployment} operates open-loop at 25 Hz: rather than using real encoders, synthetic joint observations are generated by a learned, per-joint servo MLP network driven by the policy's actions, ensuring reliable estimated servo feedback enters the joint-state observation channels. Of 60 dimensions, only 6 come from real sensors (10\%): IMU angular velocity~[3] and projected gravity~[3] via a complementary filter. Joint effort and base linear velocity are zeroed. The primary tuning parameter is the hardware scale (set to 0.75 in practice), a multiplicative factor applied to servo commands.

\section{Results}
\label{sec:results}
\begin{figure*}[!b]
\centering
\includegraphics[width=7.0in]{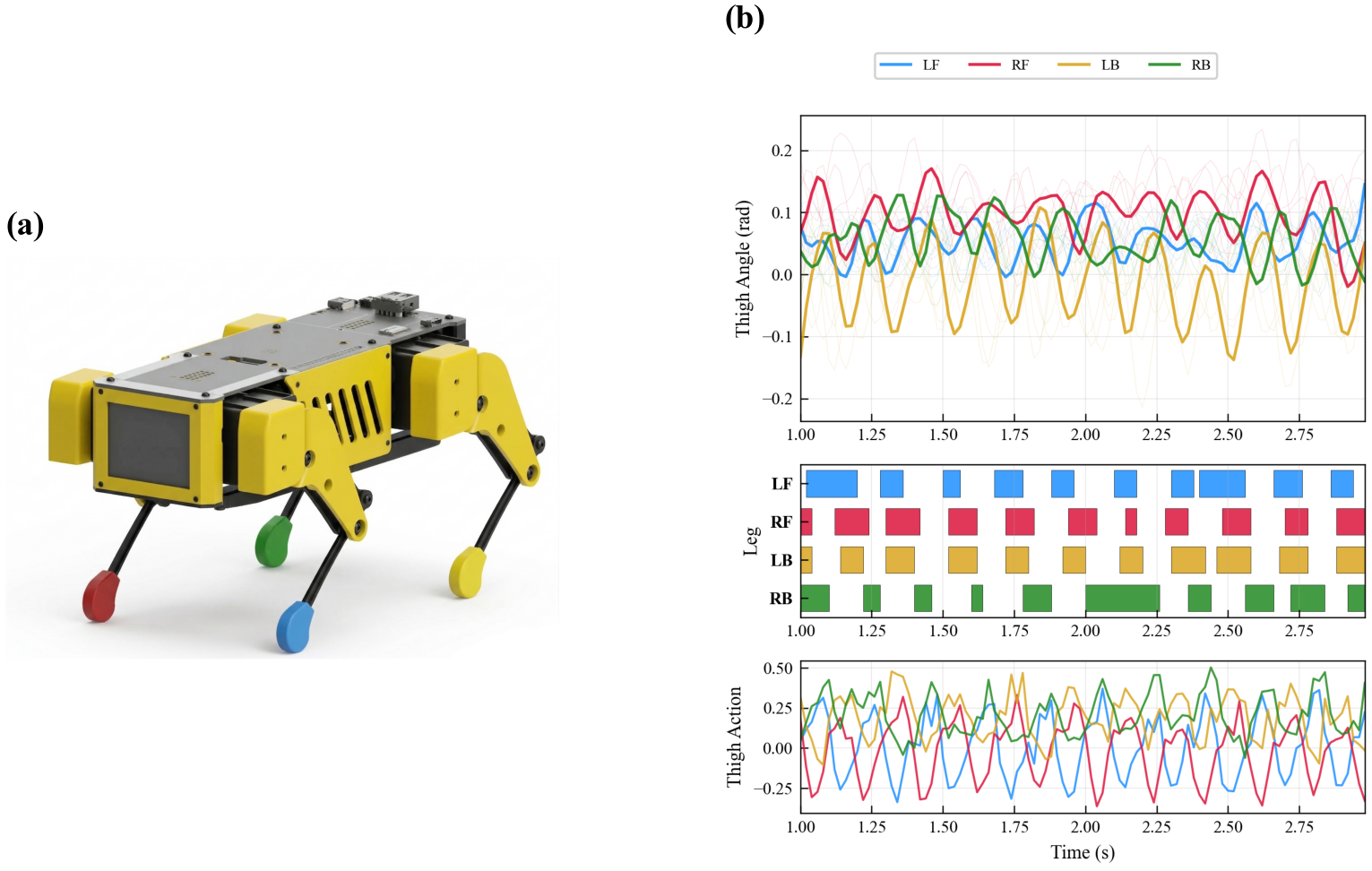}
\caption{LSTM emergent gait. (a) Mini Pupper~2. (b) Top: thigh angle traces for the four legs. Bold curves show one rollout, faint curves show 6 independent environments, which converge to consistent apparent CPG oscillations. Middle: gait phase diagram showing diagonal stance/swing coordination (trot-like pattern). Bottom: Thigh action for four legs over several gait cycles. All joints oscillate at approximately 2.8\,Hz on hardware.}
\label{fig:lstm_gait}
\end{figure*}


We evaluated each of the four candidate architectures, MLP, LSTM, GRU, and Transformer, along three axes: (1) training convergence and curriculum requirements, (2) deployment complexity on real hardware, and (3) robustness to delay perturbation.

\subsection{Emergent Central Pattern Generator}
\label{sec:results_cpg}

The delay and feedback constraints of low-cost hardware closely resemble those of biological sensorimotor systems, which manage long feedback latencies through generating autonomous rhythms in the spinal cord. Under these same constraints, we found that the LSTM converged to precisely this class of solution: a central pattern generator (CPG) despite being trained end-to-end with PPO on a standard locomotion reward, with no oscillator structure imposed. On hardware, where sensor feedback is unreliable, the learned policy exhibited the defining features of a CPG: (1) the hidden state intrinsically generated a rhythmic motor pattern  (2) the pattern did not require sensory confirmation, (3) real IMU input modulated the pattern for balance, and (4) the fundamental gait rhythm was autonomous.

Gait waveform analysis of the deployed LSTM revealed the CPG structure quantitatively (Fig. 4). Sinusoidal fit ($R^2$ to a single-frequency sine) across all 12 joints yields mean $R^2 = 0.40$ in simulation (complex multi-frequency waveform with corrections, foot placement, contact responses) and mean $R^2 = 0.77$ on hardware. The Butterworth low-pass filter, 25\,Hz control rate, and open-loop deployment removes the high-frequency corrective behaviors; and the CPG oscillation remained robust  at 2.8\,Hz, uniform across all joints.

\begin{figure*}[!t]
\centering
\includegraphics[width=7.0in]{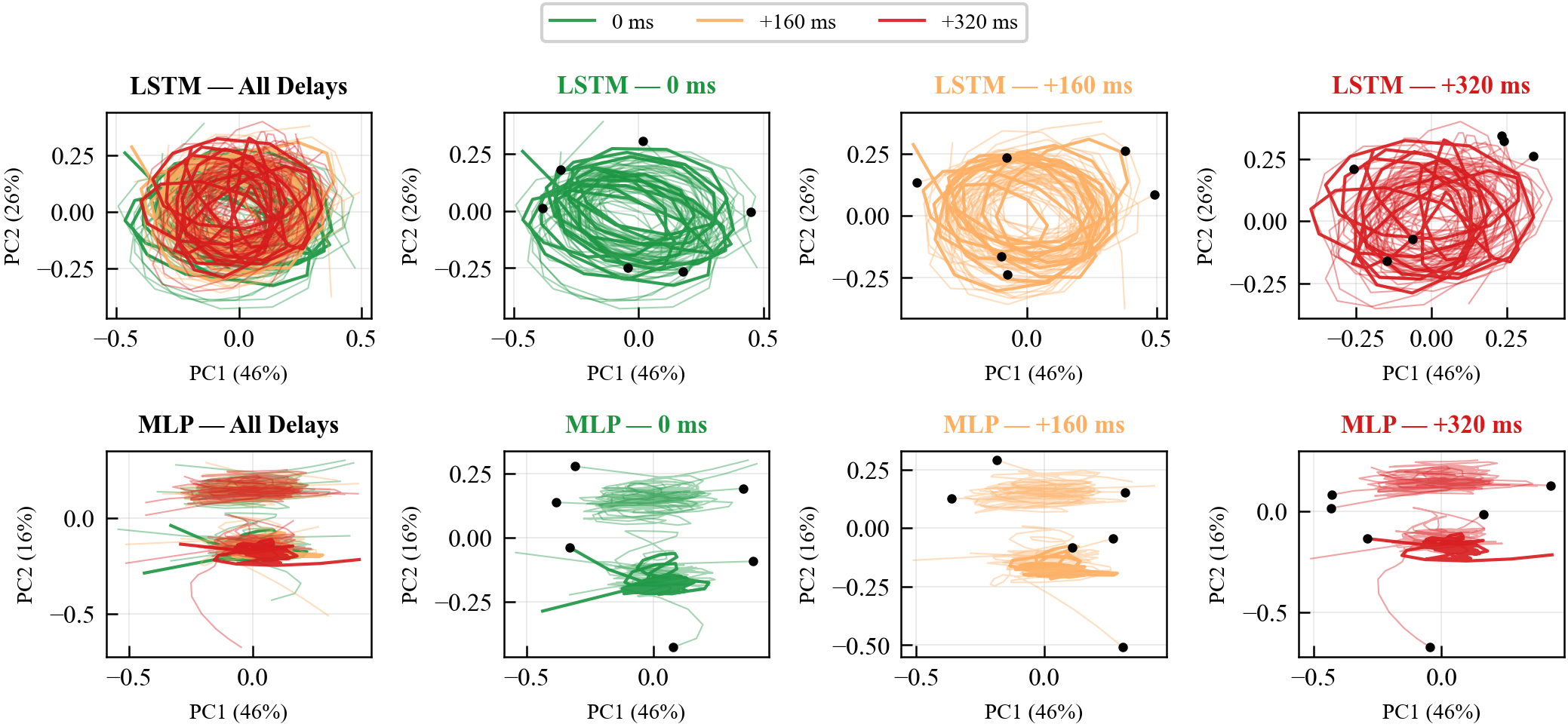}
\caption{Limit-cycle attractor analysis under diagnostic delay perturbation. PCA projections computed jointly across all delay conditions for consistent axes. Action-space attractors (12-dim $\to$ PC1--PC2): LSTM orbits are preserved at +320\,ms extra delay; MLP orbits collapse into flat horizontal bands, losing oscillatory structure. \textbf{Top:} Internal state attractors (128-dim $\to$ PC1--PC2): LSTM cell state traces a robust elliptical orbit across 0, +160, +320\,ms delay (60.4\% variance in 3 PCs). \textbf{Bottom:} MLP hidden-layer activations distort and migrate in PC space under delay (56.9\% variance, less structured). The LSTM possesses a true limit-cycle attractor (autonomous CPG); the MLP is a driven system shaped by current input.}
\label{fig:cpg_attractor}
\end{figure*}

The learned gait in Fig. 4 does not cleanly decompose into categorically separable gaits (walk, trot, gallop) as described in classical CPG literature~\cite{malmir2023, londonoalvarez2026}. Instead, the policy produces a trot-like pattern with continuous modulation. The diagonal limb pairs swing approximately in phase, but the waveform contains corrective components in simulation that are smoothed on hardware. The attractor dynamics (Fig.~\ref{fig:cpg_attractor}) reveal a stable limit cycle in both cell state and action space, but the orbit is elliptical rather than circular, suggesting asymmetric dynamics in the learned oscillator that may reflect hardware constraints such as the URDF joint limits and asymmetric leg loading. 

The property of feedback-independence that we observed is a defining signature of biological CPGs: vertebrate spinal circuits often maintain gait patterns after deafferentation, the removal of sensory feedback, and our open-loop LSTM reproduces exactly this behavior. The parallel extends to modulation as well: in vertebrates, vestibular input from the inner ear adjusts the rhythm for balance, a role played in our system by the IMU \cite{ijspeert2008}.

The same modulation pathway extends to terrain. In simulation, the learned CPG generalizes to out-of-distribution terrains, such as hills or troughs without retraining. The same gait rhythm adapts via the IMU gravity vector, which rotates on inclines. This terrain generalization further supports the CPG interpretation. This shows that the emergent CPG is robust to terrain, inclines, and the original delay constraint. 

\subsection{Robustness to Observation Delay: LSTM vs.\ MLP}
\label{sec:results_delay}

\begin{figure}[!t]
\centering
\includegraphics[width=3.4in]{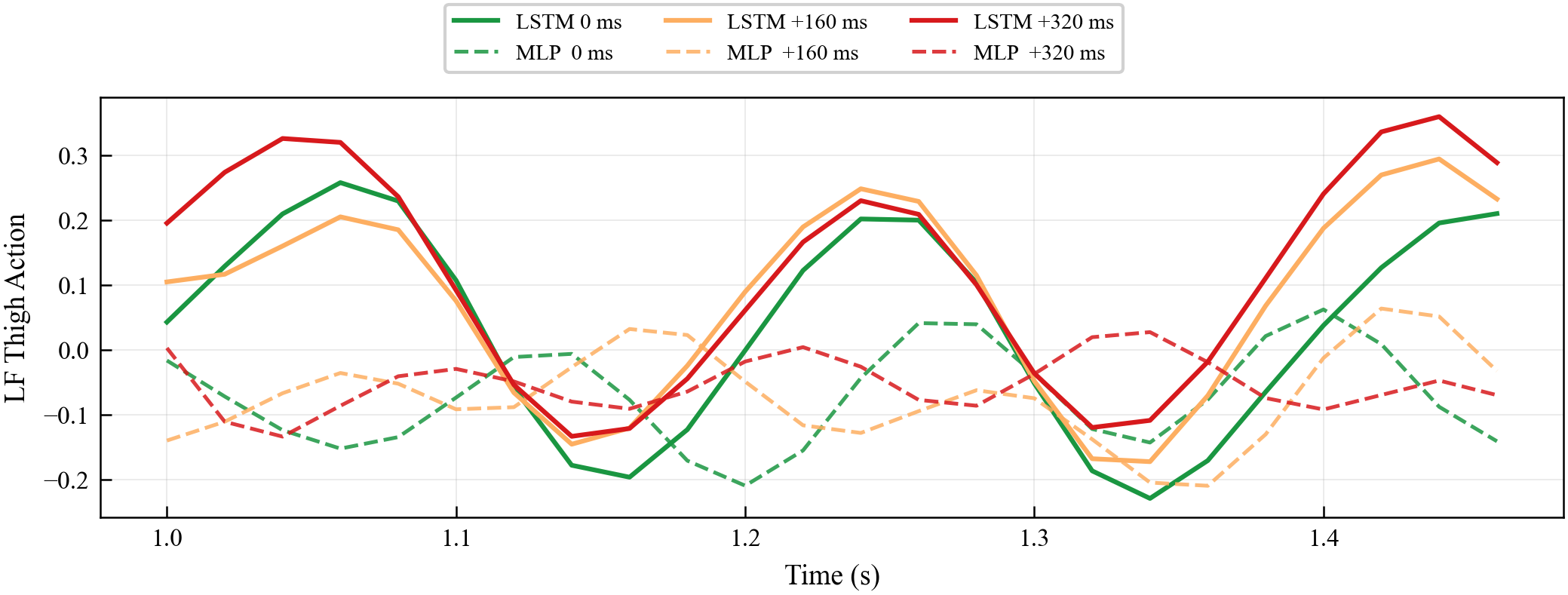}
\caption{Delay robustness: LF thigh action under 0, +160, +320\,ms extra observation delay (on top of training's 76\,ms baseline). LSTM (solid) maintains amplitude and waveform across all conditions ($\text{amp} = 0.49$ at +320\,ms). MLP (dashed) amplitude collapses to 0.16 at +320\,ms Steady-state window $t = 1.0$--$1.46$\,s, post-transient.}
\label{fig:delay_robustness}
\end{figure}

Both policies were evaluated in open-loop on their own training-environment observations with extra delay injected on joint-state channels (indices 12--48; Fig.~6). The LSTM maintained consistent 5.3\,Hz gait across all tested delay conditions, with consistent amplitude (-0.30--0.30\,rad) and low jitter. The MLP shows erratic frequency variation (6.6--7.5\,Hz), lower amplitude commands (-0.1--0.1\,rad), and higher baseline jitter. Both policies remain deployable at half the nominal command rate, within the hardware's tracking bandwidth (Fig.~2).

This robustness difference is explained by the attractor analysis (Fig.~5). In PCA projections of the action space, the LSTM traces elliptical orbits that persist across all delay conditions, while the MLP orbits collapse into flat horizontal bands as the delay increases, lacking oscillatory structure. The LSTM demonstrates a limit-cycle attractor; a CPG, while the MLP is highly input dependent.




\begin{figure*}[!t]
\centering
\subfloat[]{%
  \includegraphics[width=3.4in]{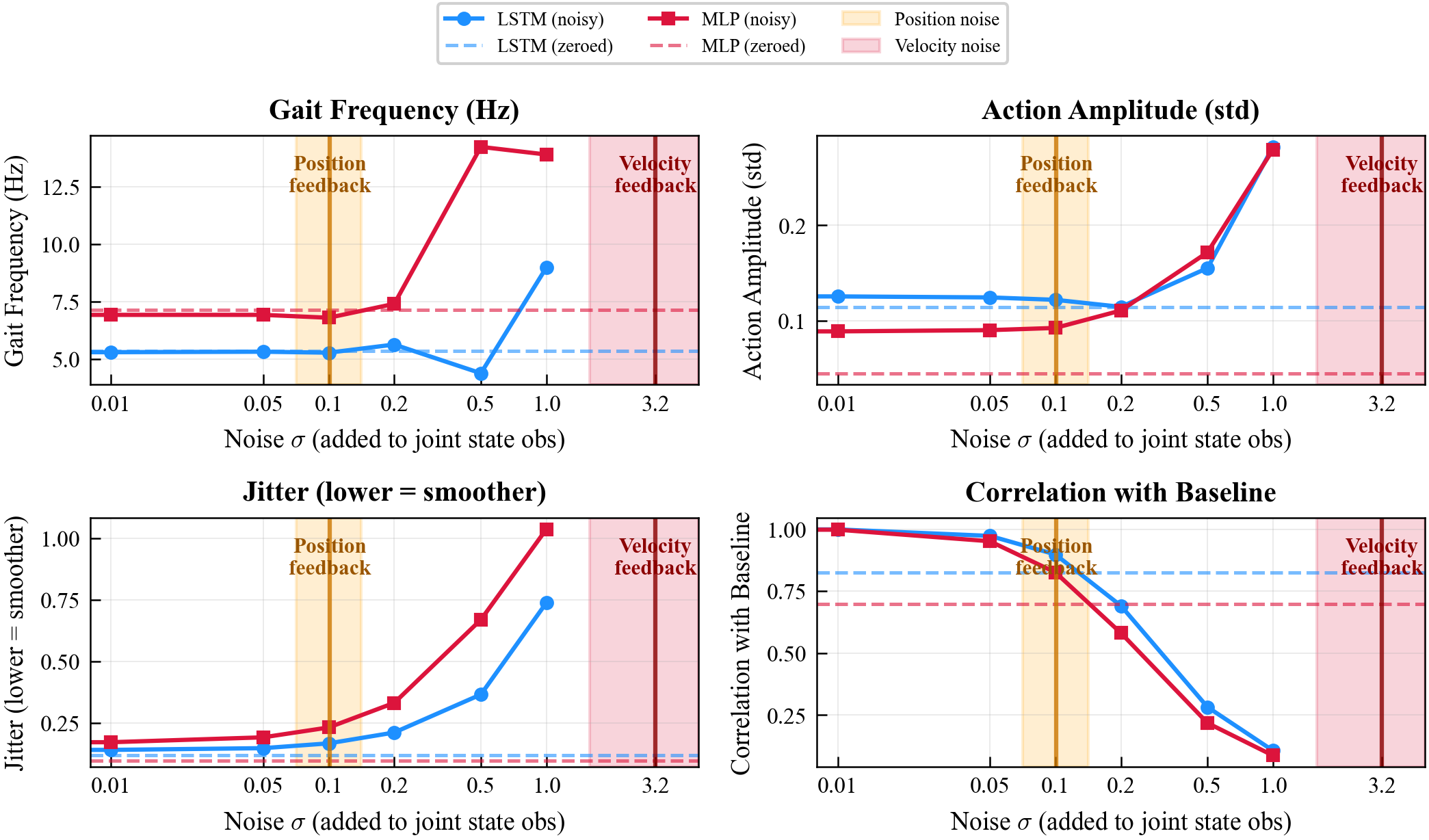}%
  \label{fig:obs_ablation}%
}
\hfill
\subfloat[]{%
  \includegraphics[width=3.4in]{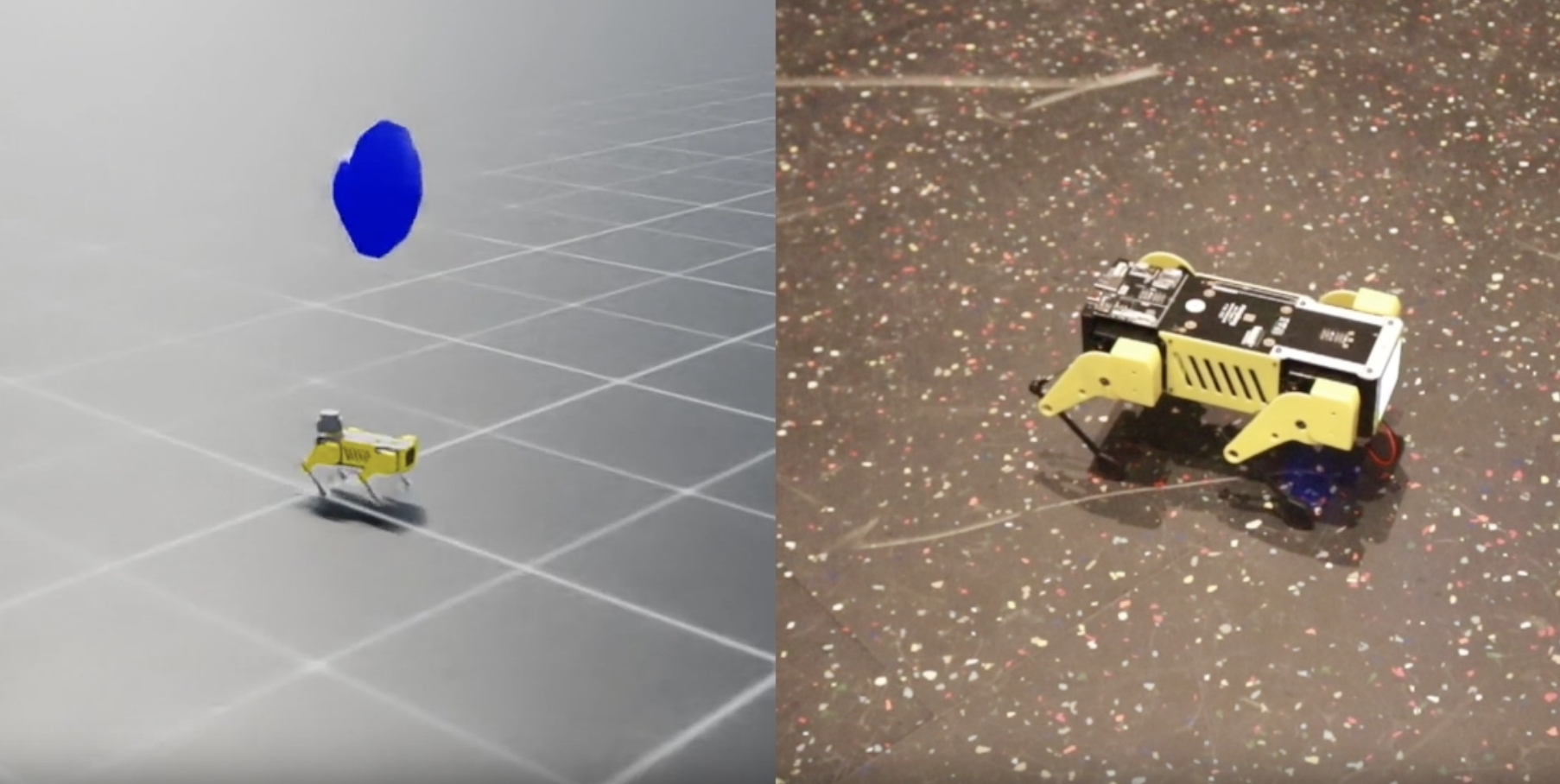}%
  \label{fig:sim_real}%
}
\caption{Observation ablation and sim-to-real deployment.
(a)~Noise vs.\ zeroed joint-state ablation showing the crossover at
$\sigma \approx 0.2$\,rad. Observation ablation: gait correlation (relative to full-observation
baseline) under Gaussian noise injected into joint-state observations
(dims 12--48) vs.\ zeroed joint state. Both LSTM (corr $= 0.826$ when zeroed)
and MLP (corr $= 0.698$) cross over at $\sigma \approx 0.2$\,rad: noise
beyond this level degrades the gait more than having no joint state at all.
Measured hardware noise levels (position $\sigma = 0.101$\,rad, velocity
$\sigma = 3.17$\,rad/s from finite differencing at 25\,Hz) are shown as
vertical bands. (b)~Sim and real Mini Pupper~2 executing
the LSTM policy open-loop. The same trained network, deployed open-loop with
hardware scale as the sole tuning parameter, produces directionally compliant
locomotion in both domains. Walking behavior is isolated and congruent, with
error accumulation primarily a result of executing commanded directions over
time. The robot demonstrates robustness for autonomous tasks, given that it
can linearly combine learned directional primitives for following or tracking
without explicit kinematic rules.}
\label{fig:ablation_and_simreal}
\end{figure*}

\subsection{Sim-to-Real Deployment}
\label{sec:results_demo}

To determine whether the policy benefits from real joint-state feedback, we conducted an observation ablation: we swept Gaussian noise of increasing standard deviation $\sigma$ into the joint-state observation channels and compared the resulting gait against zeroing those channels entirely (Fig. 7a). The zeroed condition sets the performance floor of a policy that ignores joint state altogether.

We found that both architectures exhibit a crossover near $\sigma \approx 0.2$: below this level, noisy real feedback outperforms no feedback, but above it, the joint-state signal is corrupted enough that zeroing the channels preserves the gait better than with noisy feedback. In addition, we found that the LSTM degrades more gracefully than the MLP across every metric: as the noise rises, the MLP's gait frequency destabilizes sharply and its jitter increases faster, whereas the LSTM holds a steadier rhythm and remains smoother. This indicated that the CPG rhythm of the LSTM is internally generated rather than driven by noisy observations, generating a more robust gait.

We then compared each channel's measured hardware noise to this crossover point.  Position feedback, measured at $\sigma = 0.101$ rad, falls just below the crossover and remains marginally useful. Velocity feedback, however, obtained by finite-differencing position at 25 Hz, carries $\sigma = 3.17$ rad/s, well above the crossover. On this platform it is therefore the velocity channel specifically that corrupts the observation loop, which is why we zero the velocity (and effort) channels and deploy open-loop rather than trusting the hardware's derived feedback.

Hardware deployment also revealed a property invisible in simulation in that the real servos diverged substantially from the analytic PD dynamics used in training. The servo response differs by joint and by gait phase, so a single global servo model is insufficient. Thus, each joint requires its own stateless predictor to supply accurate observations to the policy (Fig. 8).

The LSTM policy deployed open-loop with a single command gain of $0.60$--$0.75$ with a Butterworth low-pass filter at 8\,Hz on action output. Directional compliance was confirmed in all commanded directions (forward, backward, strafe, yaw) in both simulation and on the real Mini Pupper 2. Forward walking exhibited a slight clockwise arc due to hardware hip asymmetry and noisy angular feedback from the IMU gyroscope, consistent across both MLP and LSTM deployments and thus were attributable to the mechanical platform rather than the policy.

A learned per-joint MLP servo model, trained on 3 loops of open-loop playback data (390 samples per joint), replaced the analytic PD simulation for observation generation. The learned servo model reduced the joint prediction error from 0.1--1.8\,rad (PD sim) to 0.003--0.03\,rad, allowing the individual hardware servo position to have smooth and sustainable locomotion sequences. 

The LSTM-MLP deployment (Fig. 3) was thus the optimal two-layer architecture: the LSTM policy handles long-horizon temporal tasks such as delay compensation and gait generation while the learned MLP servo model handles short-horizon servo dynamics prediction. An LSTM-LSTM configuration, recurrent policy with recurrent servo model, fails for two compounding reasons. First, the servo LSTM overfits on scarce data (390 training samples), causing predicted joint positions to drift from reality. Second, because both networks share the same command signal, servo prediction errors corrupt the policy's observation of its own prior actions, generating degraded actions that further worsen servo predictions. The stateless MLP servo breaks this feedback loop because a prediction error at step t does not persist into step t+1 and the network resets. This suggests that the correct architecture for each layer of control in deployment depends on whether autoregressive rollout stability is required, not on whether the underlying process is temporal (see Extended Data, Fig. EXT1 for four-way comparison). 

\section{Discussion}
\label{sec:discussion}

The field of deep learning in robotics is young. The work of Levine~\cite{levine2016} and Schulman~\cite{schulman2017} enabled a departure from model-based control and hand-engineered state machines to end-to-end training and deployment of physical robot control. It is relatively straightforward to reproduce prior locomotion results in simulation. However, it becomes prohibitively expensive to close the sim-to-real gap on hardware. We are of the opinion that lowering the barrier for entry and democratizing contributions to the field is important given its novelty and potential. We show that deploying learned policies on low-cost hardware can lower this boundary, and that the sim-to-real bottleneck is addressable by architectural inductive bias rather than hardware fidelity. Here, we have used biological insights to address the noise and latency endemic to low-cost actuators by reframing the learning problem as a POMDP that can be learned by a time-aware network. We show (1)~time-aware networks, when given similar biological constraints, converge to a CPG as many biological systems have; (2)~these CPGs are robust to OOD latency and terrains; and (3)~the rhythmicity of the network is sustainable without actuator feedback and worsens with observation noise (Fig. 7). 
We show that quadrupedal locomotion on highly delayed hardware requires two levels of abstraction: cell-state adaptation of the LSTM for long-horizon tasks and the stateless feedback from a learned servo MLP to accurately feed the varying latency of each servo to the control policy. The stateless MLP servo is the correct choice in our low-data regime. We nonetheless hypothesize that an LSTM servo model would surpass it given thousands more samples per joint, at which point hidden-state drift would become negligible relative to the MLP's function approximation error. On our platform, collecting such data requires extended open-loop playback sessions that risk mechanical damage. Notably, this dual-layer recurrent failure mode may only become apparent in embodied deployment, where autoregressive drift compounds through the physical system in ways that purely simulated evaluations could miss. 
We demonstrate that the GRU's update gate, which applies a nonlinear transform at every step, cannot maintain the pure shift-register representation required for delay compensation, and that transformers, while powerful, are not strictly necessary for this task. Taken together, our results indicate a strategy for overcoming longstanding resource constraints on low-cost hardware: architecture selection should match the time horizon for each control layer, not simply the temporal nature of the underlying process. Moreover,  these results suggest that training under large transport delay induces CPG-like solutions in time-aware networks, which may generalize beyond quadrupedal locomotion to any rhythmic control task subject to significant transport latency.

\section{Conclusion}
\label{sec:conclusion}

We presented a complete sim-to-real pipeline for quadrupedal locomotion on a \$300 robot with a 76\,ms transport delay, comparing MLP, LSTM, GRU, and Transformer architectures under identical training conditions with a calibrated delayed PD actuator model. The LSTM produces an emergent CPG that deploys open-loop with a single tuning parameter, while the MLP requires a hand-tuned synthetic PD bridge with six or more parameters. The GRU fails (lacking the cell state pathway for delay compensation), and the Transformer is impractical for training or deploying on low-cost hardware. PCA attractor analysis and observation ablation experiments provide mechanistic explanations for these outcomes, grounded in the POMDP structure imposed by transport delay on cost-constrained hardware. On platforms where the sim-to-real gap is dominated by communication latency rather than model fidelity, recurrent architectures with linear memory pathways offer a principled path to reducing deployment engineering. More fundamentally, high-delay hardware appears to induce CPG solutions in time-aware networks; suggesting that the biological strategy of autonomous rhythm generation may be a general solution for cost-constrained embodied control.

\section*{Acknowledgments}
The authors thank the Baccus lab at Stanford University and the Ölveczky lab at Harvard University for discussions on biological central pattern generators, and the Isaac Lab and rsl\_rl communities for open-source training infrastructure. We thank Shuya Gong for figure edits, Kyrstyn Ong for control theory suggestions on subsampling techniques in simulation and hardware, and Shenghua Liu for helpful discussions. We thank the NSF GRFP and Stanford Bio-X Bowes Fellowship for enabling this research.

{\appendix[Extended Data]

\begin{figure*}[!t]
\centering
\includegraphics[width=7.0in]{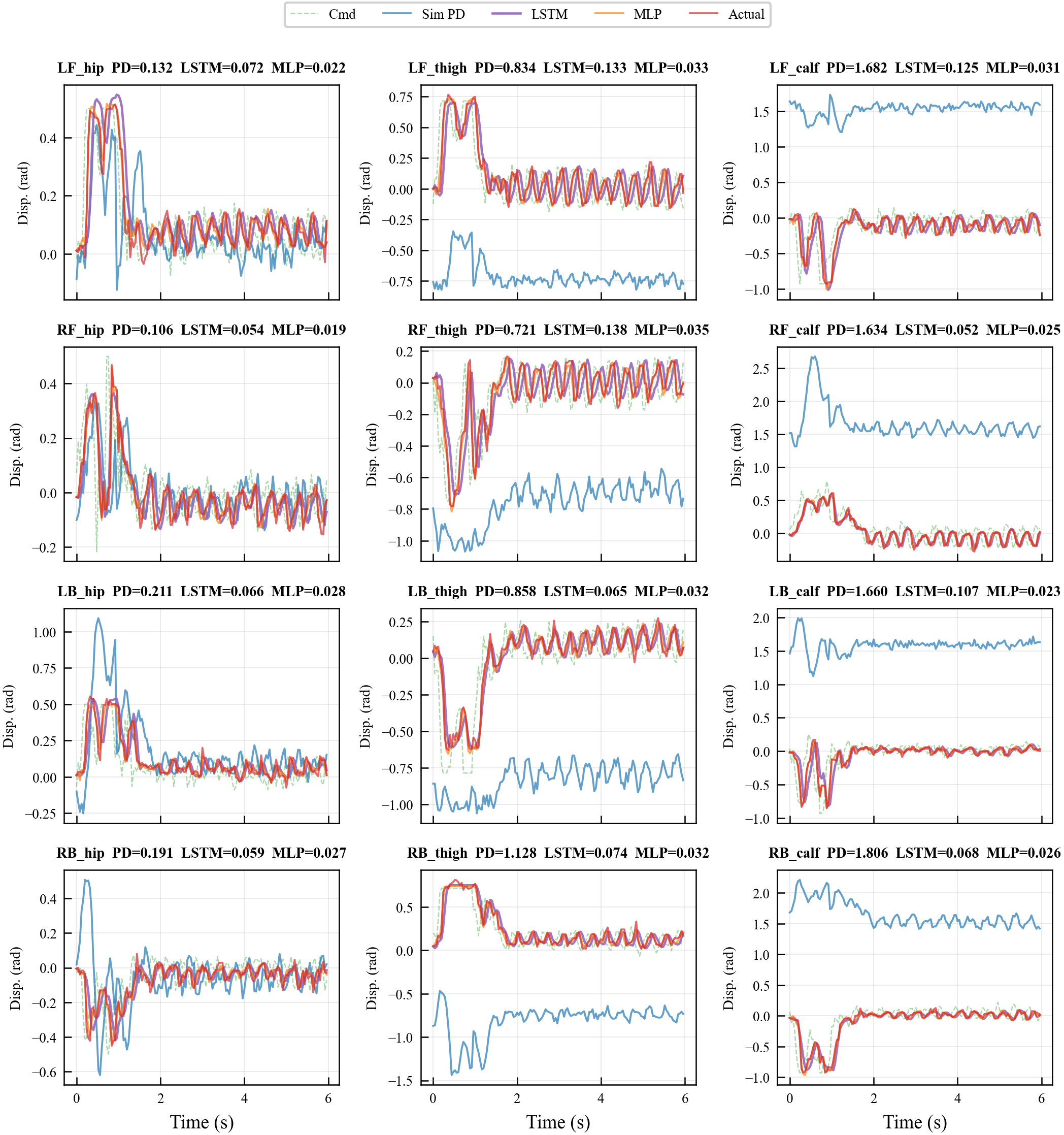}
\caption{Four-way observation source comparison for LSTM deployment: Sim PD (analytic DelayedPDActuator), learned LSTM servo model, learned MLP servo model, and actual real servo response. Per-joint RMSE shown in subplot titles. The MLP servo model (RMSE 0.003--0.03\,rad) closely tracks actual servo dynamics across all 12 joints, while the analytic PD sim (RMSE 0.1--1.8\,rad) diverges significantly. The LSTM servo model achieves better single-step prediction but accumulates drift during autoregressive rollout (mean RMSE 0.070 vs.\ MLP 0.028), confirming that the stateless MLP architecture is correct for the data-scarce, short-horizon servo modeling task. This motivates the two-layer deployment stack: LSTM policy (long-horizon temporal reasoning) with MLP servo model (short-horizon rollout stability).}
\label{fig:ext1_four_way}
\end{figure*}
}

\end{document}